\documentclass[mathematics,article,submit,pdftex,moreauthors]{Definitions/mdpi} 
\firstpage{1} 
\pubvolume{1}
\issuenum{1}
\articlenumber{0}
\pubyear{2022}
\copyrightyear{2022}
\datereceived{} 
\dateaccepted{} 
\datepublished{} 
\hreflink{https://doi.org/} 

\usepackage{fontenc,inputenc,calc,indentfirst,fancyhdr,graphicx,epstopdf, lastpage,ifthen,lineno,float,amsmath,setspace,enumitem,mathpazo,booktabs,titlesec,etoolbox,tabto,xcolor,soul,multirow,microtype,tikz,totcount,changepage,attrib,upgreek,amsthm,hyphenat,natbib,footmisc, url,geometry,newfloat,url, geometry, newfloat, caption, pifont, ulem}

\makeatletter
\let\c@lofdepth\relax
\let\c@lotdepth\relax
\makeatother
\usepackage{subfigure}
\graphicspath{{./image/}}

\Title{Group-Mix SAM: Lightweight Solution for Industrial Assembly Line Applications}

\TitleCitation{Title}

\Author{Liang~Wu $^{1,2}$, X.-G.~Ma $^{1,2,3,4}$*}

\AuthorNames{Liang~Wu, X.-G.~Ma}

\AuthorCitation{Wu, Liang, X.-G.~Ma}

\address{%
$^{1}$ \quad Faculty of Robot Science and Engineering, Northeastern University, Shenyang, China\\
$^{2}$ \quad Foshan Graduate School, Northeastern University, Foshan, China\\
$^{3}$ \quad College of Information Science and Engineering, Northeastern University, Shenyang, China\\
$^{4}$ \quad The State Key Laboratory of Synthetical Automation for Process Industries, Northeastern University, Shenyang, China\\
}

\corres{Correspondence: maxg@mail.neu.edu.cn}

\abstract{
Since the advent of the Segment Anything Model(SAM) approximately one year ago, it has engendered significant academic interest and has spawned a large number of investigations and publications from various perspectives. However, the deployment of SAM in practical assembly line scenarios has yet to materialize due to its large image encoder, which weighs in at an imposing 632M. In this study, we have replaced the heavyweight image encoder with a lightweight one, thereby enabling the deployment of SAM in practical assembly line scenarios. Specifically, we have employed decoupled distillation to train the encoder of MobileSAM in a resource-limited setting. The entire knowledge distillation experiment can be completed in a single day on a single RTX 4090. 
The resulting lightweight SAM, referred to as Group-Mix SAM, had 37.63\% (2.16M) fewer parameters and 42.5\% (15614.7M) fewer floating-point operations compared to MobileSAM. However, on our constructed industrial dataset, MALSD, its mIoU was only marginally lower than that of MobileSAM, at 0.615. Finally, we conducted a comprehensive comparative experiment to demonstrate the superiority of Group-Mix SAM in the industrial domain. With its exceptional performance, our Group-Mix SAM is more suitable for practical assembly line applications.
}

\preto{\abstractkeywords}{\nolinenumbers} 
\begin{document}
\section{Introduction}
ChatGPT\cite{zhang2023one} thoroughly transformed the NLP field and made groundbreaking progress in large-scale models. In the field of computer vision, the role of SAM was undoubtedly prominent. SAM achieved state-of-the-art performance in zero-shot instance segmentation tasks and achieved great success. SAM was compatible with other models, enabling advanced visual applications such as text-guided segmentation and fine-grained image editing. Despite these advantages, it was difficult to deploy the SAM model in actual assembly lines. The reason is that using SAM to perform segmentation tasks in actual deployment would result in unbearable computing and memory costs. SAM can be viewed as two parts: the ViT-based image encoder and the prompt-guided mask decoder. The parameter size of the ViT-based image encoder was two orders of magnitude larger than that of the prompt-guided mask decoder, with sizes of 632M and 3.87M\cite{zhang2306faster}, respectively. The mask decoder was lightweight, so the fundamental reason why SAM cannot be deployed in actual assembly lines is the large parameter size of the image encoder.

In actual assembly lines, it was the edge computers that were responsible for running the algorithms. The purchase of these edge computers was usually limited by price, so there were issues such as low memory and weak computing power. Therefore, to deploy and use in actual assembly line scenarios, we needed to replace the heavyweight image encoder with a lightweight image encoder to reduce the size of SAM.

In this paper, we addressed the issue of MobileSAM's inability to be deployed in assembly lines due to the lack of computing power in edge computers and excessive memory usage. We proposed replacing the original ViT-T structure in MobileSAM's encoder with a smaller image encoder structure called Groupmixformer\cite{ge2023advancing}, which achieved excellent results. We named the resulting model Group-Mix SAM. As stated in Kirillov et al.\cite{kirillov2023segment}, training a SAM with ViT-H image encoder required 68 hours on 256 A100 GPUs. Directly training with a small image encoder was time-consuming, laborious, and the results were not necessarily good. Therefore, we turned to knowledge distillation techniques. Decoupled distillation, which directly extracted the small image encoder from the original SAM's ViT-H without relying on a combined decoder, was superior in terms of both time and effectiveness compared to semi-coupled (freezing the mask decoder and optimizing the image encoder from the mask layer) and coupled distillation (optimizing the image encoder directly from the mask layer)\cite{zhang2306faster}. Therefore, we chose decoupled distillation as the method for knowledge distillation. Concretely, it transferred knowledge from the ViT-T based structure to the Groupmixformer with a smaller image encoder.

\section{Related works}
\subsection{Vision Transformers}
The Vision Transformers (ViTs) exhibited astonishing performance in visual applications through the use of multi-head self-attention (MHSA) mechanism, which effectively captured global dependencies. Compared to CNN, ViT models demonstrated higher upper limits and generalization, but also came with practical deployment issues due to their high parameter counts. Therefore, many papers and projects on ViT focused on efficient deployment, such as LeViT\cite{graham2021levit}, MobileViT\cite{mehta2021mobilevit}, Tiny-ViT\cite{wu2022tinyvit}, EfficientViT\cite{liu2023efficientvit}, GroupMixFormer\cite{ge2023advancing}, etc. Designing efficient ViT structures could construct efficient SAM.
\subsection{Segment Anything Model}
SAM\cite{kirillov2023segment} was widely used in visual tasks as it could segment any object in an image based on point or box prompt. Its zero-shot generalization was learned based on the SA-1B dataset, and it exhibited outstanding performance in various visual tasks. Additionally, SAM was also applied in fields such as medical image segmentation and transparent object detection. Due to its extensive practical applications, the actual production line deployment of SAM has received increasing attention. Recent studies proposed some strategies to reduce the computational cost of SAM, such as EfficientSAM\cite{xiong2023efficientsam}, which introduced a pre-training framework named SAMI that utilized the mask image of SAM, edge SAM\cite{zhou2023edgesam}, which extracted the original ViT-based SAM image encoder into a purely CNN-based architecture, and MobileSAM\cite{zhang2306faster}, which obtained a lightweight image encoder through decoupling distillation to replace the original bulky encoder of SAM. Our focus was on achieving the actual production line deployment of SAM.
\subsection{Knowledge Distillation}
Knowledge distillation (KD) was a method used to transfer the ability of a model. Depending on the method of transfer, it can be divided into two main directions: target distillation (logits method distillation) and feature distillation. The most classic paper in target distillation is \cite{hinton2015distilling}, which proposed concepts such as soft-target and temperature. In contrast, feature distillation involves the student learning the intermediate layer features of the teacher network structure. The earliest paper to propose this method was FitNets\cite{romero2014fitnets}, which forced the student's intermediate layer to approximate the corresponding intermediate layer features of the teacher, achieving the goal of feature distillation. Most of the research on knowledge distillation was focused on tasks such as classification, semantic segmentation, and object detection\cite{guan2020differentiable}\cite{xie2020self}\cite{zhang2020distilling}\cite{shu2021channel}\cite{zhang2020improve}.
\section{Methodology}
In this section, several key methods used in the paper will be introduced. First, we will introduce the image encoder we finally selected: Groupmixformer and its principles and advantages, and then introduce the distillation method we used: decoupled distillation and its implementation principle.
\subsection{Groupmixformer}
Groupmixformer\cite{ge2023advancing} proposed Group-Mix Attention(GMA) as an advanced alternative to traditional self-attention. Compared with the popular multi-head self-attention which models only the correlation between each token, GMA could utilize group aggregators to capture the correlations between token-to-token, token-to-group, and group-to-group simultaneously. Groupmixformer achieved excellent performance in downstream tasks such as image classification, object detection, and semantic segmentation.

In section 4.1, we conducted a thorough comparative experiment on the Encoders that are suitable for industrial production line scenes, and Groupmixformer performed the best. Therefore, we selected it as the student model for knowledge distillation.
\subsection{Distillation}
In knowledge distillation, the teacher and student architectures served as the carriers for knowledge transfer. In other words, the quality of knowledge transfer from the teacher to the student depended on how the network structures of the teacher and student were designed. Based on previous experience, we selected the decoupled distillation method and used the encoder, vit-t, in MobileSAM\cite{zhang2306faster} as the teacher model and Groupmixformer as the student model. For the distillation framework, we adopted nanosam\footnote{https://github.com/NVIDIA-AI-IOT/nanosam}, which used NVIDIA TensorRT to greatly accelerate the inference speed of the teacher model during the distillation process. In addition, it used the feature maps obtained by the teacher and student through forward inference to calculate the final loss, and this simple mechanism also sped up the distillation process. Therefore, our complete distillation process could be low-cost and reproduced on a single RTX 4090 in less than a day.

\section{Experiments and Analysis}
\begin{table}[!t]
\caption{Training and Testing Existing Image Encoders on the Midea Assembly Line Dataset. \textbf{Bold} in the table represents the best, and \uline{underlined} represents the second best.}
\begin{tabular}{cccccc}
\hline
\textbf{Encoder Name}   & \textbf{dataset} & \textbf{train loss} & \textbf{train acc} & \textbf{val loss} & \textbf{val acc} \\ \hline
Deepvit\cite{zhou2021deepvit}                 & MALS             & 0.3945              & 0.8608             & 0.4091            & 0.8562           \\
NesT\cite{gewaltig2007nest}                    & MALS             & 0.4032              & 0.8608             & 0.4099            & 0.8571           \\
RegionVit\cite{chen2021regionvit}               & MALS             & 0.2121              & 0.8956             & 0.3001               & 0.8801             \\
Vit\cite{dosovitskiy2020image}                     & MALS             & 0.2082              & 0.9098             & 0.2934            & 0.8819           \\
CCT\cite{hassani2104escaping}                     & MALS             & 0.1381              & 0.9446             & 0.2248            & 0.8909           \\	
CrossViT\cite{chen2021crossvit}                & MALS             & 0.0506              & 0.9858             & 0.3730             & 0.8978           \\
Token-to-Token ViT\cite{yuan2021tokens}      & MALS             & 0.1284              & 0.9562             & 0.2305            & 0.8988           \\
PiT\cite{heo2021rethinking}                     & MALS             & 0.2220               & 0.9098             & 0.2661            & 0.8988           \\
CaiT\cite{touvron2021going}                    & MALS             & 0.1095              & 0.9665             & 0.285             & 0.9127           \\
CvT\cite{wu2021cvt}                     & MALS             & \uline{0.0057}        & \uline{0.9920}       & 0.3573            & 0.9266           \\
LeViT\cite{graham2021levit}                   & MALS             & 0.0402              & 0.9910              & \uline{0.1903}      & \uline{0.9345}     \\
\textbf{Groupmixformer\cite{ge2023advancing}} & MALS             & \textbf{0.0388}     & \textbf{0.9927}     & \textbf{0.1542}   & \textbf{0.9673}  \\ \hline
\end{tabular}
\label{table:1}
\end{table}
In this section, we will introduce the experimental process, results, and analysis of the paper. Firstly, we will introduce the experimental process and results of selecting an encoder suitable for practical industrial assembly line scenarios. Then, we will introduce the dataset and experimental settings used for knowledge distillation and present the results of the distillation experiments. Finally, we will compare the performance of Group-Mix SAM and MobileSAM.
\begin{figure}[!h]
    \centering
    \includegraphics[width=0.6\linewidth]{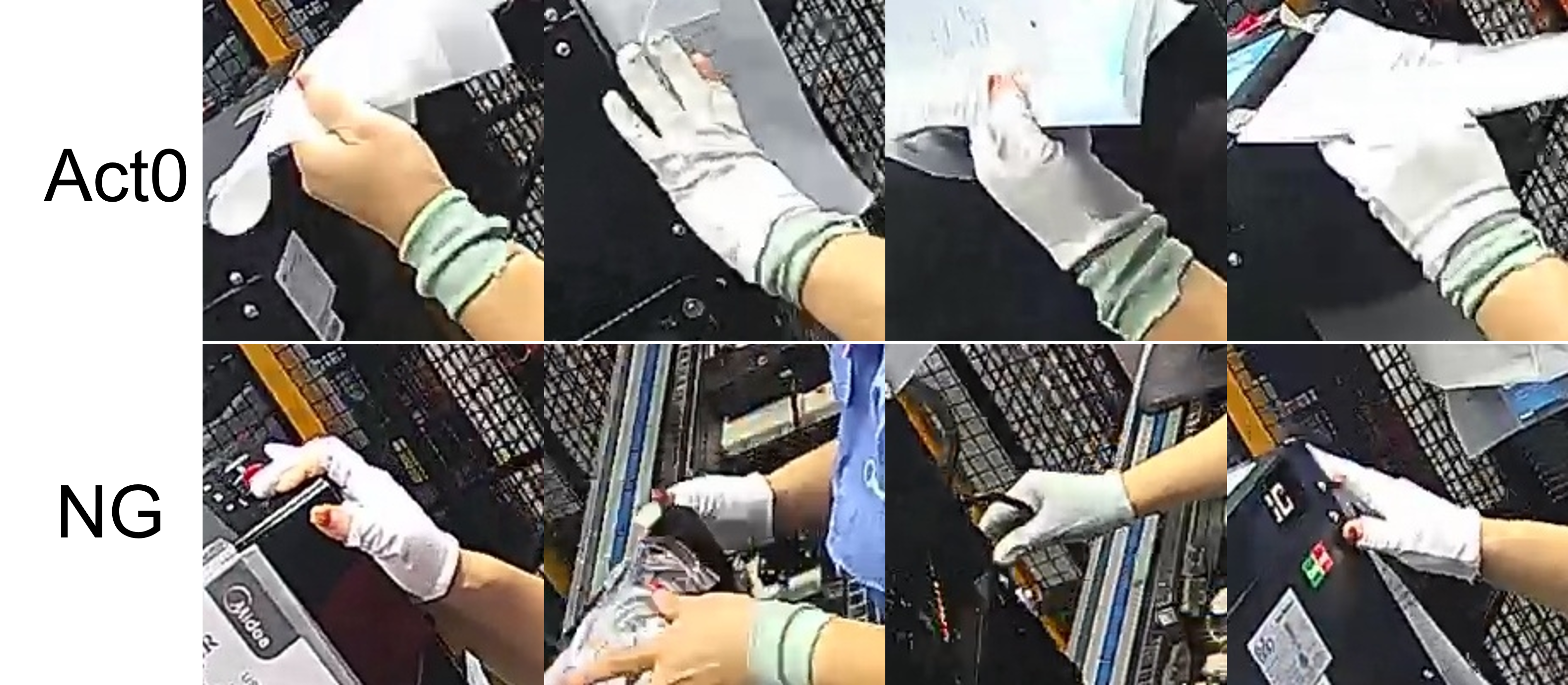}
    \caption{Example of Midea Assembly Line Scenario Dataset(MALSD)}
    \label{fig:1}
\end{figure}
\subsection{Experiments on selecting image encoders}
We needed to select an encoder that was suitable for actual industrial assembly line scenarios. To do this, we created the Midea Assembly Line Scenario Dataset (MALSD) by collecting video data from Midea's actual assembly line. As shown in Fig \ref{fig:1}, MALSD had only two categories: Act0 and NG. Act0 referred to images of workers holding product instructions, while ng referred to other miscellaneous images. The training set had 533 and 668 images for Act0 and NG, respectively, and the test set had 144 and 238 images for Act0 and NG, respectively. As shown in Tab \ref{table:1}, we trained and tested existing state-of-the-art image encoders on the MALSD. We observed that groupmixformer\cite{ge2023advancing} performed the best on MALSD. It outperformed the second-best model, LeViT, by 3.28\% on the validation set. Therefore, we concluded that Groupmixformer\cite{ge2023advancing} is better suited for our industrial scenario.

\subsection{Experimental setup for knowledge distillation}
We selected Nanosam as the distillation framework. The COCO 2017\cite{lin2014microsoft} dataset was used, with unlabeled images for training and labeled images for testing. The image size fed to the student during distillation was 1024, with a batch size of 8, the learning rate of 3e-4, and 13 epochs uniformly set for all experiments. The loss function used was Huber. To reduce the time required for distillation, we accelerated the teacher model using TensorRT. As a result, the entire distillation process was completed on a single RTX 4090 in less than a day.
\begin{figure}[!h]
    \centering
    \includegraphics[width=0.6\linewidth]{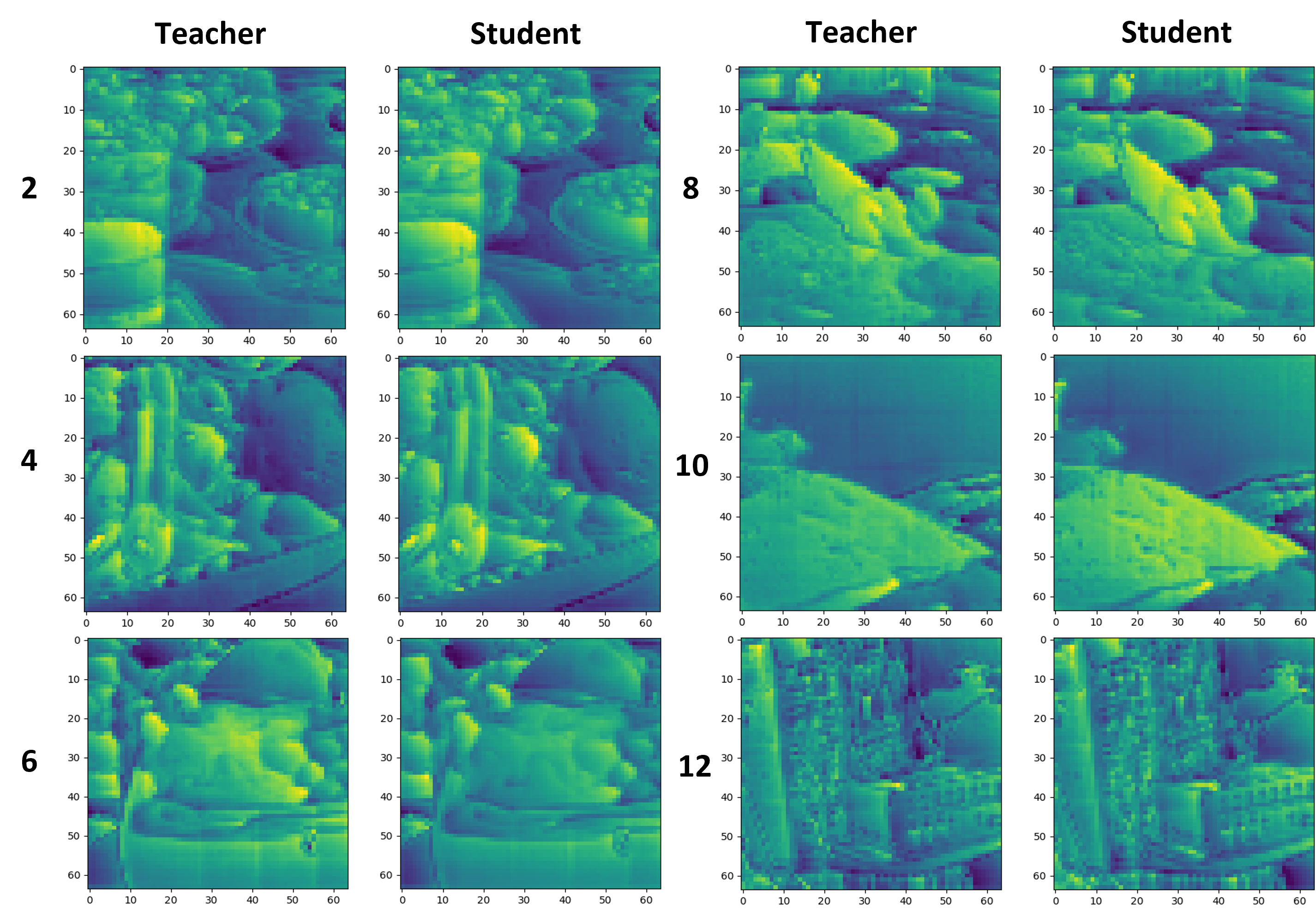}
    \caption{Comparison of characteristics of teachers and students during knowledge distillation. The numbers in the picture represent epoch, the feature size is (1,256,64,64), and the first dimension (64*64) is visualized here}
    \label{fig:2}
\end{figure}
\subsection{Experimental results of knowledge distillation}
The results were shown as depicted in Tab \ref{table:2}. We selected MobileSAM's encoder, Vit-T, as the teacher model, which had a parameter count of 5.74M and a floating-point operation count of 36742.79M. It performed at 72.8\% on the coco dataset and showed excellent performance in smaller models. We used three different structures, Resnet18, Vit-T with the same structure as the teacher model, and Groupmixfromer (GMF), as the student models for knowledge distillation. We then tested the internal structure of Groupmixfromer, such as reducing the number of stages from four to three, reducing the serial depths of a layer, adjusting the learning rate, etc. Finally, we determined that the Groupmixfromer with the last row structure of [2,2,8,2] was the most suitable for our industrial scenario. Although its accuracy was 0.3\% lower than Vit-T, it had 37.63\%(2.16M) fewer parameters and 42.5\%(15614.7M) fewer floating-point operations, making it deployable in an industrial assembly line scenario. The feature comparison pictures during training was shown in Fig \ref{fig:2}.     
\begin{table}[!t]
\caption{The experimental record table for knowledge distillation is presented below. \textbf{T/S} indicates whether the model is a teacher or a student, \textbf{Structure} represents the model structure after modification, \textbf{Lr} denotes the learning rate, Epoch denotes the number of learning rounds, \textbf{Last loss} represents the final loss of knowledge distillation, and \textbf{accuracy} was tested on the COCO 2017 dataset. $\dag$ represents that the input is uniformly (1, 3, 1024, 1024), and the hardware is a single RTX 4090. $\star$ represents the teacher model of this experiment.}
\begin{tabular}{ccccccccc}
\hline
Model & T/S & Structure    & Lr     & Epoch & Last loss        & Accuracy       & Params         & Flops$\dag$          \\ \hline
Vit-T$\star$     & T   & -            & -      & -     & -                & 0.716          & 5.74M          & 36742.79 M          \\ 
Resnet18       & S   & -            & 0.0003 & 13    & 0.00050          & 0.693          & 14.78 M        & 45623.54 M          \\
Vit-T          & S   & -            & 0.0003 & 13    & \textbf{0.00023} & \textbf{0.708} & 5.74M          & 36742.79 M          \\
GMF            & S   & [3, 3, 12,4] & 0.0003 & 13    & 0.00030          & \textbf{0.708} & 5.63M          & 32435.40 M          \\
GMF            & S   & [3, 3, 12,4] & 0.0030 & 13    & 0.00033          & 0.701          & 5.63M          & 32435.40 M          \\
GMF            & S   & [3, 3, 12,4] & 0.0010 & 13    & 0.00029          & \textbf{0.708} & 5.63M          & 32435.40 M          \\
GMF            & S   & [3, 3, 12]   & 0.0003 & 13    & 0.00034          & 0.703          & 4.31M          & 26808.09M           \\
GMF            & S   & [3, 3, 4,4]  & 0.0003 & 13    & 0.00036          & 0.701          & \textbf{3.03M}          & 21407.53 M          \\
GMF            & S   & [2, 2, 6, 4] & 0.0003 & 13    & 0.00034          & 0.702          & 3.58M & \textbf{21128.09 M} \\
GMF            & S   & [2, 2, 8, 2] & 0.0003 & 13    & 0.00034          & 0.703          & 3.58M & \textbf{21128.09 M} \\
GMF            & S   & [2, 2, 8, 2] & 0.0010 & 13    & 0.00033          & 0.705          & 3.58M & \textbf{21128.09 M} \\ \hline
\end{tabular}
\label{table:2}
\end{table}

\subsection{Performance of Group-Mix SAM}
\begin{figure}[!h]
    \centering
    \includegraphics[width=0.8\linewidth]{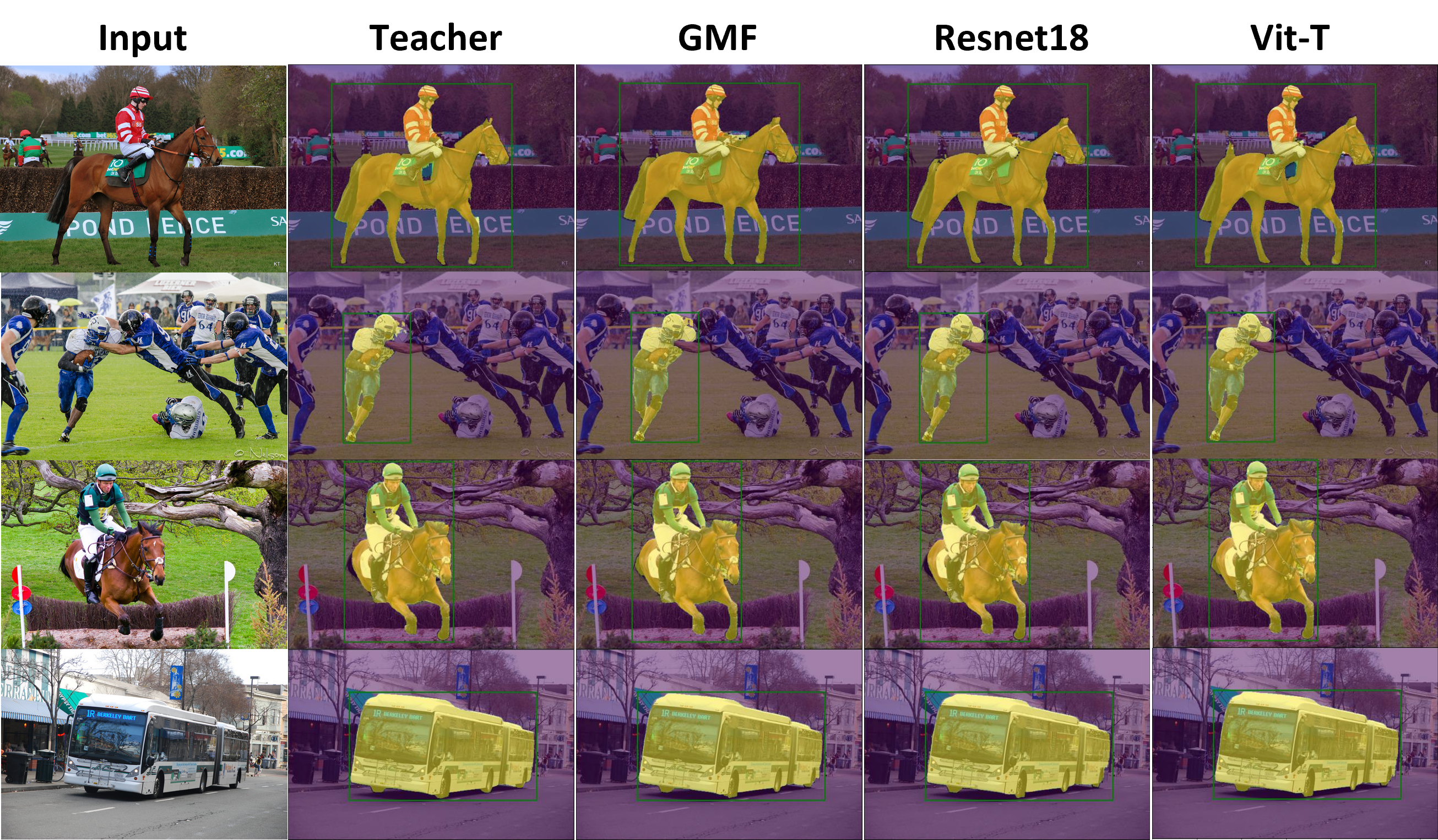}
    \caption{Performance comparison between teacher models and student models, namely GMF, Resnet18, and Vit-T on coco dataset. Viewing the comparison in color and zoom-in is recommended for better visualization.}
    \label{fig:3}
\end{figure}

The image encoder GMF, which was distilled and suitable for industrial assembly line scenarios, was used in image segmentation, resulting in Group-Mix SAM. We qualitatively compared the distilled Teacher model, GMF, Vit-T, and Resnet18 backbones, which replaced the encoder in Mobile SAM, using the COCO public dataset. We used bounding boxes as the prompt in the experiment. As shown in Fig \ref{fig:3}, Group-Mix SAM seems to perform better in some details, such as the horse's tail and back in the first and third rows and the Teacher model, Resnet18, and Vit-T all have some flaws.

\begin{figure}[!h]
    \centering
    \includegraphics[width=0.8\linewidth]{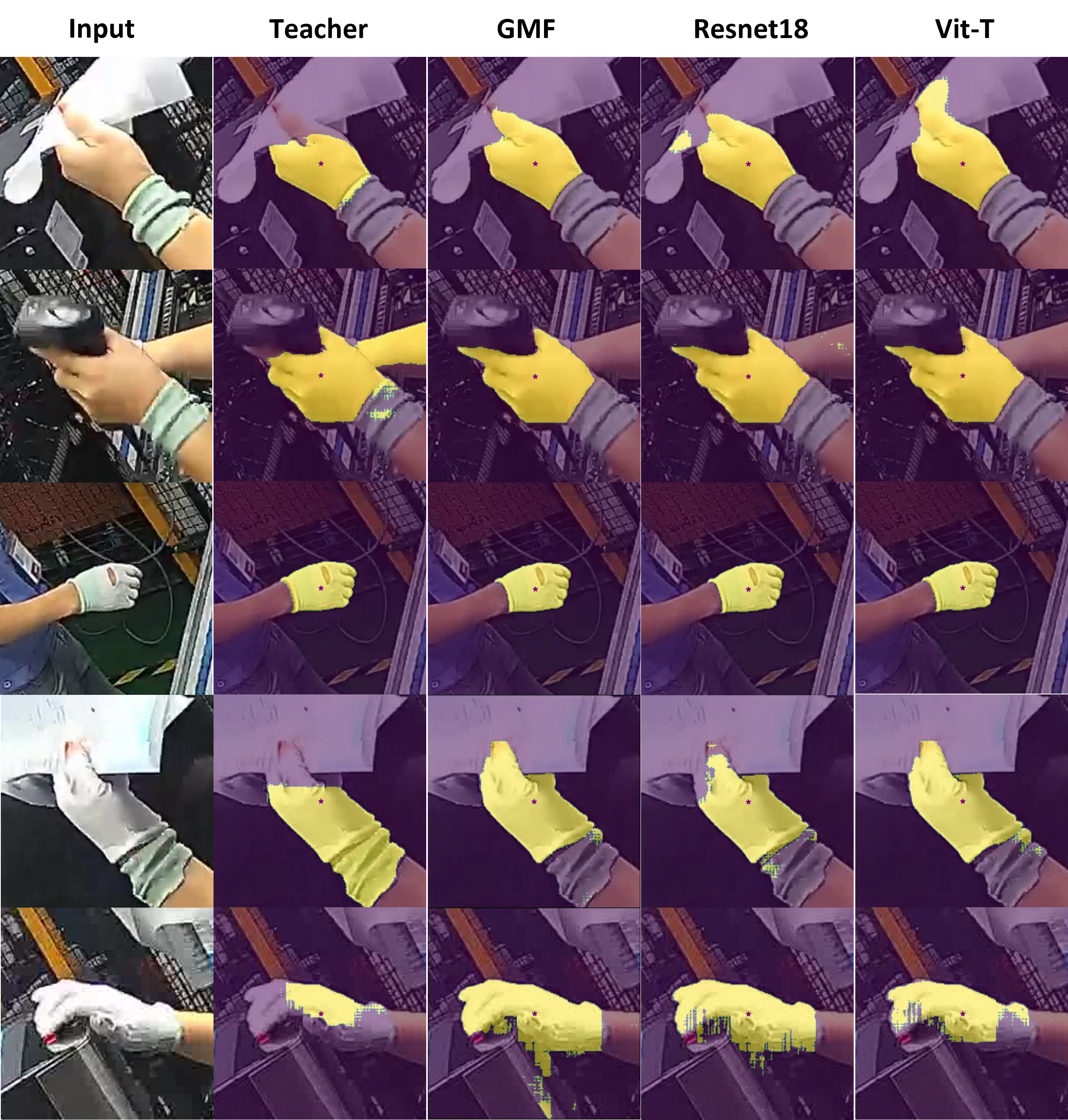}
    \caption{Performance comparison between teacher models and student models, namely GMF, Resnet18, and Vit-T on MALSD dataset. Viewing the comparison in color and zoom-in is recommended for better visualization.}
    \label{fig:4}
\end{figure}

We then repeated the above experiment and conducted a qualitative comparison on our own constructed dataset: MALSD, this time using the center point of the worker's hand as the prompt. As shown in Fig \ref{fig:4}, despite having the smallest parameter and floating-point operation counts, GMF still performed well. For example, in the first and fourth rows where workers hold instruction manuals, and in the second row where workers hold barcode scanners, GMF was able to accurately and clearly segment the hands compared to the other three methods.

Finally, we conducted a detailed test of Group-Mix SAM using the above four methods on MALSD, and the results are shown in Tab \ref{table:3}.
We can see that in MALSD, mIoU is lower than measured in conventional datasets (MobileSAM can reach more than 70\% in conventional datasets) because there are many interferences and noises in industrial datasets. And although the encoder used by Group-Mix SAM is distilled from MobileSAM (37.63\%(2.16M) fewer parameters and 42.5\%(15614.7M) fewer floating-point operations), there is not much difference in the industrial dataset.

\begin{table}[!t]
\caption{mIoU comparison. Assuming that the predicted mask from the original SAM is ground truth, a higher mIoU indicates a better performance.}
\begin{tabular}{ccccc}
\\ \hline
              & Encoder        & Dataset & Prompt & mIoU  \\ \hline
MobileSAM     & Vit-T          & MALSD   & Point  & 0.623 \\ 
Group-Mix SAM & Groupmixformer & MALSD   & Point  & 0.615 \\ \hline
\end{tabular}
\label{table:3}
\end{table}
\section{Conclusion}
We selected an encoder suitable for practical industrial assembly line scenarios from our self-constructed industrial dataset, MALSD. Subsequently, we employed decoupled distillation to distill the encoder of MobileSAM into the Gounpmixformer architecture. By replacing the heavyweight image encoder with a lightweight one, we enabled the deployment of SAM in practical assembly line scenarios. With its exceptional performance, our Group-Mix SAM is better suited for practical assembly line applications.

\authorcontributions{}

\funding{The research was supported by the startup fund of Fashan graduate school, Northeastern University (No. 200076421002), the Natural Science Foundation of Guangdong Province (No. 2020A1515011170), and the 2020 Li Ka Shing Foundation Cross-Disciplinary Research Grant (No. 2020LKSFG05D).}

\dataavailability{The dataset and source code generated during and/or analysed during the current study are available from the corresponding author on reasonable request.}


\conflictsofinterest{No potential conflict of interest was reported by the authors. } 

\begin{adjustwidth}{-\extralength}{0cm}

\bibliography{references}
\bibstyle{mdpi}


%



\end{adjustwidth}
\end{document}